\pdfoutput=1

\documentclass[11pt]{article}

\usepackage[final]{acl}
\usepackage{graphicx}
\usepackage{amsmath}
\usepackage{amssymb}
\usepackage{booktabs}
\usepackage{float}
\usepackage{caption}
\usepackage{colortbl}
\usepackage{booktabs}
\usepackage{marvosym}
\usepackage{multicol}
\usepackage{multirow}
\usepackage{lipsum}
\usepackage{times}
\usepackage{latexsym}

\usepackage[T1]{fontenc}

\usepackage[utf8]{inputenc}

\usepackage{inconsolata}
\usepackage{enumitem}

\usepackage{graphicx}
\usepackage[most]{tcolorbox}
\newtcolorbox{blackpromptbox}[1][]{%
breakable,
colback=gray!10,
colframe=black,
fontupper=\small,
left=0.5mm, right=0.5mm, top=1mm, bottom=1mm,
boxrule=0.8pt,
sharp corners,
title={Prompt},
fonttitle=\bfseries,
#1 
}

\title{VC4VG: A Comprehensive Video Caption Optimizing Strategy for Text-to-Video Generation}
\title{VC4VG: Optimizing Video Captions for Text-to-Video Generation}

\author{
 \textbf{Yang Du\textsuperscript{1}}\thanks{Equal contribution.},
 \textbf{Zhuoran Lin\textsuperscript{2}}$^*$,
 \textbf{Kaiqiang Song\textsuperscript{2}}$^*$,
 \textbf{Biao Wang\textsuperscript{2}},
 \textbf{Zhicheng Zheng\textsuperscript{2}},
 \\
 \textbf{Tiezheng Ge\textsuperscript{2}},
 \textbf{Bo Zheng\textsuperscript{2}},
 \textbf{Qin Jin \textsuperscript{1}}\thanks{Corresponding author: \href{qjin@ruc.edu.cn}{qjin@ruc.edu.cn}}
\\
 \textsuperscript{1}School of Information, Renmin University of China,
 \\
 \textsuperscript{2}Taobao \& Tmall Group of Alibaba,
}

\begin{document}
\maketitle
\begin{abstract}
Recent advances in text-to-video (T2V) generation highlight the critical role of high-quality video-text pairs in training models capable of producing coherent and instruction-aligned videos. However, strategies for optimizing video captions specifically for T2V training remain underexplored. In this paper, we introduce \textbf{VC4VG} (\textbf{V}ideo \textbf{C}aptioning for \textbf{V}ideo \textbf{G}eneration), a comprehensive caption optimization framework tailored to the needs of T2V models.
We begin by analyzing caption content from a T2V perspective, decomposing the essential elements required for video reconstruction into multiple dimensions, and proposing a principled caption design methodology. To support evaluation, we construct VC4VG-Bench, a new benchmark featuring fine-grained, multi-dimensional, and necessity-graded metrics aligned with T2V-specific requirements.
Extensive T2V fine-tuning experiments demonstrate a strong correlation between improved caption quality and video generation performance, validating the effectiveness of our approach. We release all benchmark tools and code\footnote{\url{https://github.com/qyr0403/VC4VG}} to support further research.
\end{abstract}

\section{Introduction}
\label{sec:intro}

\begin{figure*}[!t]
	\centering
	\includegraphics[width=0.9\linewidth]{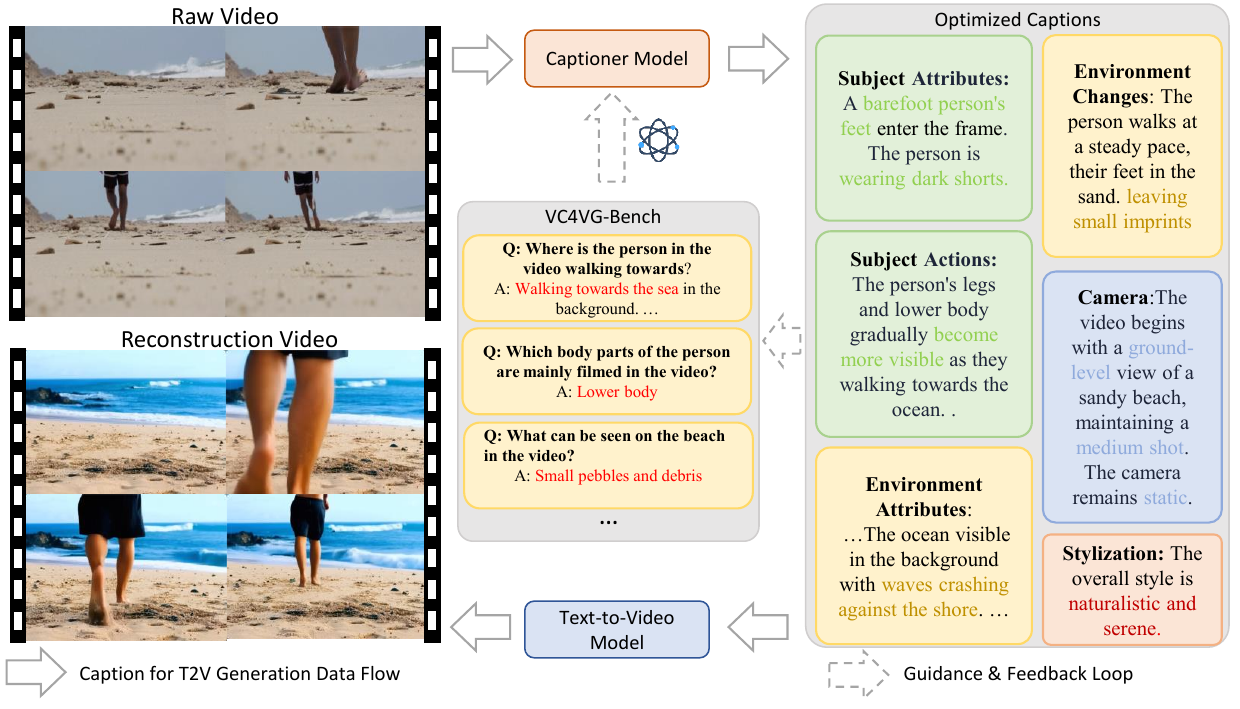}
    \caption{Overview of the video caption optimization framework for text-to-video (T2V) generation. The original video is transformed into textual descriptions via captioners. These captions are then optimized according to dimensions that we consider essential for video reconstruction and instruct by VC4VG-Bench evaluation. Finally, optimized captions are used during T2V models' training and generating videos.
    }
	\label{fig:overall}
\end{figure*}

Text-to-video (T2V) generation has witnessed rapid progress in recent years, marked by impressive systems such as Sora~\cite{openai2024sora} and Kling\cite{kuaishou2024kling}. A core driver behind these advancements is the availability of large-scale, high-quality video-caption pairs that enable T2V models to generate visually rich and instruction-aligned content. However, acquiring such high-quality video-text pairs remains a major bottleneck: although large volumes of video data are readily available online, most lack accurate textual annotations or are labeled with low-quality captions. To bridge this gap, recent large-scale datasets have increasingly relied on automated captioning powered by multimodal large language models (MLLMs)~\cite{chen2024panda70m, wang2023internvid}. 

As a result, emerging T2V systems (e.g., OpenSora~\cite{opensora}, CogVideoX~\cite{yang2024cogvideox}) and curated datasets (e.g., OpenVid~\cite{nan2024openvid}, ShareGPT4Video~\cite{chen2025sharegpt4video}, Miradata~\cite{ju2025miradata}) have adopted pseudo-caption generation as a key preprocessing step. Despite this trend, there remains a critical gap: no existing work provides a systematic caption optimization framework that aligns caption design, evaluation, and T2V training in a unified, feedback-driven loop. Meanwhile, existing video captioning benchmarks suffer from two key limitations: 1) They rely on outdated metrics (e.g., BLEU~\cite{papineni2002bleu}, CIDEr~\cite{vedantam2015cider}) designed for short and generic captions. 2) They lack evaluation protocols tailored to the specific needs of video generation tasks (e.g., AuroraCap~\cite{chai2024auroracap}, Dream-1K~\cite{ wang2024tarsier_recipestrainingevaluating}).

To address these limitations, we propose \textbf{VC4VG} (\textit{Video Captioning for Video Generation}), a comprehensive caption optimization framework specifically designed to enhance T2V training. As illustrated in Figure~\ref{fig:overall}, VC4VG consists of three key components: 

\noindent\textbf{Dimension-Aware Caption Optimization}:
From a T2V generation perspective, we analyze the core visual-linguistic elements required for video reconstruction and decompose captions into five essential dimensions:
(1) subject attributes, (2) environmental context, (3) motion dynamics, (4) camera parameters, and (5) atmospheric/stylistic elements. We hypothesize that rich and accurate coverage across these dimensions contributes directly to improve video generation performance. We therefore optimize raw captions generated by the captioner according to these dimensions. 

To investigate how dimensional optimizations enhance T2V generation relative to other caption models, and to enable efficient large-scale captioning on datasets with over 10M videos, we build a custom MLLM captioner, LLaVA-Video-Gen-7B. It builds on LLaVA-Video~\cite{zhang2024llavavideo}, augmented with Gemini 1.5 Pro~\cite{team2024gemini1.5} and temporal-sensitive data from RTime~\cite{Du2024RTime}, and supports scalable, locally deployable, high-quality caption generation.

\noindent\textbf{VC4VG-Bench --- A T2V-Generation-Oriented Benchmark}: We introduce VC4VG-Bench, a hierarchical, LLM-assisted benchmark comprising ~1,000 human-annotated Video–QA pairs. These QAs span multi-level visual content, from high-level themes to fine-grained visual details.
To measure caption effectiveness, we introduce a necessity-based hierarchy that distinguishes \textit{core} from \textit{supplementary} content for video reconstruction. This allows for automated, LLM-as-judge evaluations that align well with human assessments, enabling scalable and accurate evaluation of captioning quality from a generation-oriented perspective and providing actionable insights for model selection and data optimization in text-to-video generation.

\noindent\textbf{Closed-Loop Validation via T2V Fine-tuning}: To validate the practical utility of our framework, we fine-tune CogVideoX~\cite{yang2024cogvideox} on three versions of a 72K-sample video-caption dataset curated from OpenVid-1M~\cite{nan2024openvid}, using captions generated by different methods, including CogVLM2-Caption~\cite{yang2024cogvideox}, LLaVA-Video-7B~\cite{zhang2024llavavideo}, and our proposed LLaVA-Video-Gen-7B (served as a proof-of-concept implementation of our optimization framework).  
Quantitative results on VBench~\cite{huang2023vbench, huang2024vbench++} and MovieGenBench~\cite{polyak2024moviegencastmedia}, together with qualitative studies, show that generation quality correlates strongly with the richness and necessity alignment of caption content across our defined dimensions, validating the effectiveness of our optimization strategy.

Our main contributions are threefold: 1) We systematically decompose video captioning into five key dimensions critical to video reconstruction, providing guidance for scalable caption generation. 2) We propose a benchmark with ~1,000 human-verified QA pairs and an automated evaluation protocol tailored to T2V needs. 3) We demonstrate, through fine-tuning experiments, that improvements in caption content directly enhance video generation quality, validating our caption optimization strategy. 

\section{VC4VG}

\label{sec:benchmarks and methods}
we propose \textbf{VC4VG} (Video Captioning for Video Generation), a comprehensive caption optimization strategy tailored for enhancing T2V training. In this section, we first present caption information dimensions decomposed from the essential requirements of T2V reconstruction, accompanied by the development of LLaVA-Video-Gen, a captioner for large-scale video captioning in Section~\ref{paradigm}. We then introduce VC4VG-Bench, a novel benchmark specifically designed for video captioning from the text-to-video generation perspective in Section~\ref{cap4gen-bench}. 

\subsection{Caption Optimization}
\label{paradigm}
\label{method:dimensions}

High-quality video-caption pairs are essential for effective T2V training. We hypothesize that rich and accurate coverage across key dimensions in captions directly enhances video generation performance. To validate this, we systematically decompose video captioning into five critical dimensions ensuring comprehensive yet flexible coverage of essential content. This decomposition is grounded in a systematic analysis of the fundamental requirements of T2V generation, drawing inspiration from practices used by professional video creators. These dimensions include:

\begin{itemize}[leftmargin=*,noitemsep] 
    \item \textbf{Camera Parameter Specification}: Camera parameters capture the perspective from which the content is viewed which shapes narrative framing and viewer engagement. They critically govern text-to-video generation through three key dimensions: (1) \textit{shot size} defining subject scale relative to the frame, (2) \textit{camera angles} specifying viewpoint orientation, and (3) \textit{movement patterns} describing dynamic transitions inferred by analyzing scene context and static reference objects. Special techniques like slow motion or macro shots are explicitly annotated as \textit{shot technology} modifiers.
    \item \textbf{Subject Attributes}: A clearly defined subject serves as the semantic core of the scene and is essential for T2V models to generate meaningful, instruction-aligned content. We define subjects as the primary objects in a video, characterized by two key visual aspects: 1) basic properties such as quantity, appearance, clothing, and accessories; 2) spatial relationships among subjects, including their positions and interactions.
    \item  \textbf{Motion Dynamics}: Motion is the defining feature of video compared to static images and its accurate modeling is essential for achieving temporal coherence. We define motion dynamics through three core elements: (1) Gradual environmental changes over time, (2) Sequential actions broken down into detailed limb movements, and (3) Movement paths showing direction and position changes when subjects travel through scenes.
    \item \textbf{Environmental Contexts}: The environment defines the spatial and visual setting in which the subject appears, directly influencing lighting, composition, and physical interactions. This dimension is fundamental to building a believable world. We set environment descriptions encompass: (1) Spatiotemporal attributes (lighting conditions, weather, time-of-day), (2) Geospatial layout with object placements, and all elements are grounded in visually observable evidence without subjective interpretation.
    \item \textbf{Stylization Guidelines}: This dimension determines the final artistic rendering, influencing the overall appearance to meet user-specific stylistic preferences. We summarize high level visual aspects through: (1) Emotional ambiance conveyed via color grading and motion patterns, (2) Stylistic descriptors (e.g., anime, cyberpunk) governing rendering pipelines. These are derived from low-level visual cues rather than external semantic knowledge.
\end{itemize}

\subsubsection{LLaVA-Video-Gen:A Proof-of-Concept}
While powerful, existing MLLMs like LLaVA-Video-7B~\cite{zhang2024llavavideo} lack explicit optimization for generating the complex, instruction-driven descriptions required for high-quality T2V training. 
To reduce this gap and validate our framework, we introduce LLaVA-Video-Gen, a 7B-parameter expert captioner as a proof-of-concept for our framework. This model is developed by distilling Gemini 1.5 Pro~\cite{team2024gemini1.5}, into the more efficient LLaVA-Video-7B architecture. Our data curation and fine-tuning pipeline consists of two complementary stages.

\textbf{General-Purpose Captioning Data Curation.}
First, to enhance the foundational capability to follow complex instructions for diverse visual concepts, we curate a high-quality dataset from WebVid-10M~\cite{Bain21webvid}. Our multi-step filtering process is designed to maximize data quality and diversity. We initially select videos with durations between 5 and 15 seconds to ensure sufficient content richness while aligning with typical T2V generation lengths. To foster content diversity, we employ Qwen2VL~\cite{qwen2-VL} to extract content tags (e.g., subject, environment) for balanced sampling across different concepts. A subsequent data cleaning pipeline (in Appendix ~\ref{appendix:video_filter}) further filters this subset based on aesthetic quality and motion intensity, resulting in 200K high-quality videos. Crucially, we discard the original, often noisy WebVid captions and use Gemini 1.5 Pro to generate entirely new, detailed descriptions, ensuring high linguistic consistency and semantic depth.

\textbf{Temporal Reasoning Enhancement.}
Second, to specifically enhance the model's temporal reasoning—a known weakness in many MLLMs—we incorporate the RTime dataset~\cite{Du2024RTime}. RTime contains 21K videos featuring distinct forward and reversed semantics (e.g., "opening a door" vs. "closing a door"), each paired with manually verified short captions. We leverage these concise, high-confidence captions as contextual prompts to guide Gemini 1.5 Pro in generating long-form, temporally-aware descriptions. The resulting data, structured as (video, forward\_caption, reversed\_caption) triples, is naturally suited for Direct Preference Optimization (DPO)~\cite{rafailov2023dpo}.

\textbf{Fine-tuning.}
Using the comprehensive collection of generated captions, we fine-tune the LLaVA-Video-7B model using Low-Rank Adaptation (LoRA)~\cite{hu2022lora}. For each video, we uniformly sample 32 frames for training. The DPO-based fine-tuning on the RTime data further sharpens the model's ability to distinguish and describe temporal sequences, yielding our expert captioner, LLaVA-Video-Gen. Additional ablation studies are provided in Appendix~\ref{appendix:abalation_llava}.

\begin{figure}[!t]
	\centering
	\includegraphics[width=\linewidth]{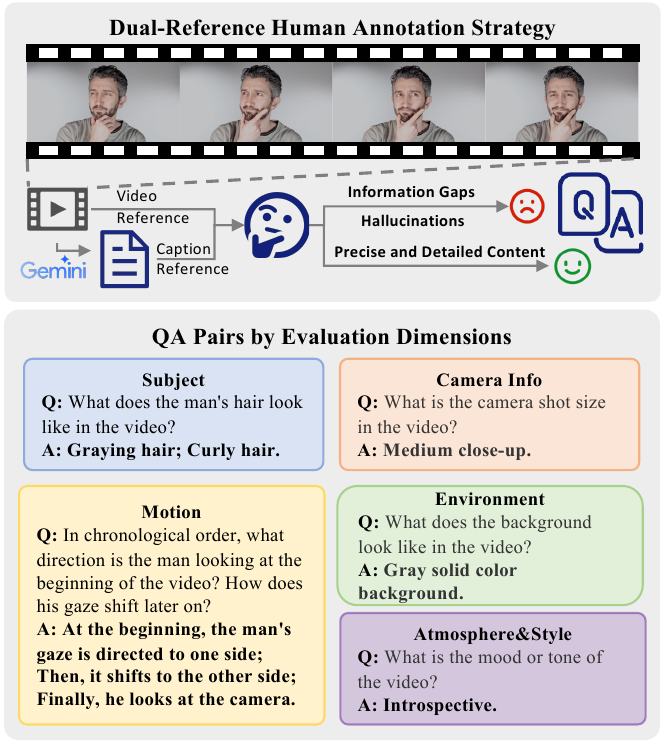}
	\caption{The core framework of evaluation QA-pairs, structured around five key assessment dimensions. Leveraging dual-reference (video content \& textual captions) enables multimodal alignment verification, effectively assisting human annotation to ensure accuracy and comprehensive coverage in evaluation QA-pairs.}
	\label{fig:cap4gen bench}
\end{figure}

\begin{figure}[!t]
	\centering
	\includegraphics[width=\linewidth]{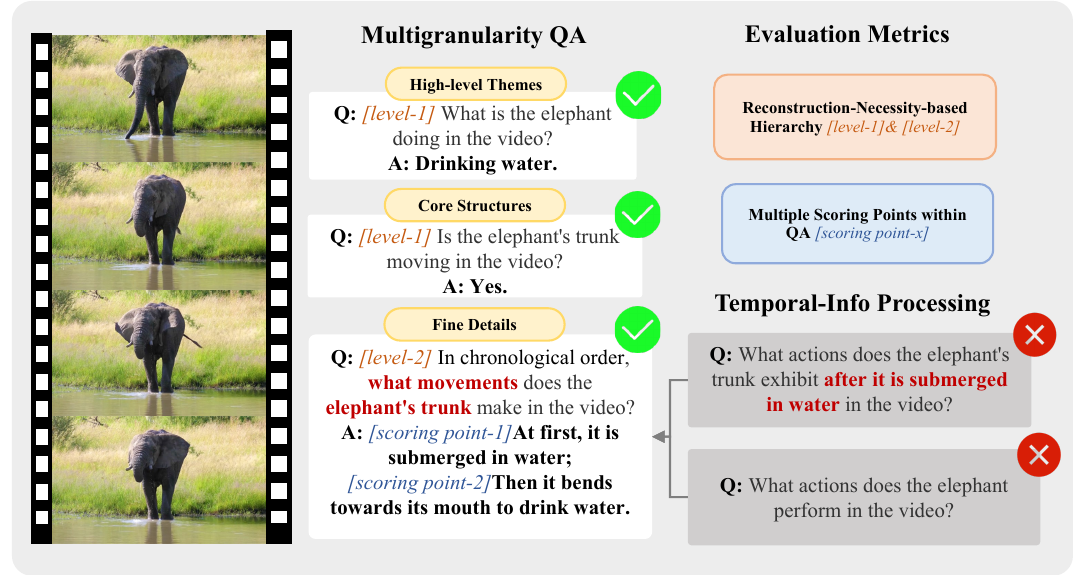}
	\caption{Illustration of the multi-granularity evaluation QA-pair system specifically designed for video generation tasks. Featuring moderate information clustering in temporal processing, the hierarchical QA-pair architecture based on reconstruction-necessity incorporates multiple scoring points to comprehensively assess caption quality in video generation tasks.
}
	\label{fig:metric1}
\end{figure}

\begin{figure}[!h]
	\centering
	\includegraphics[width=\linewidth]{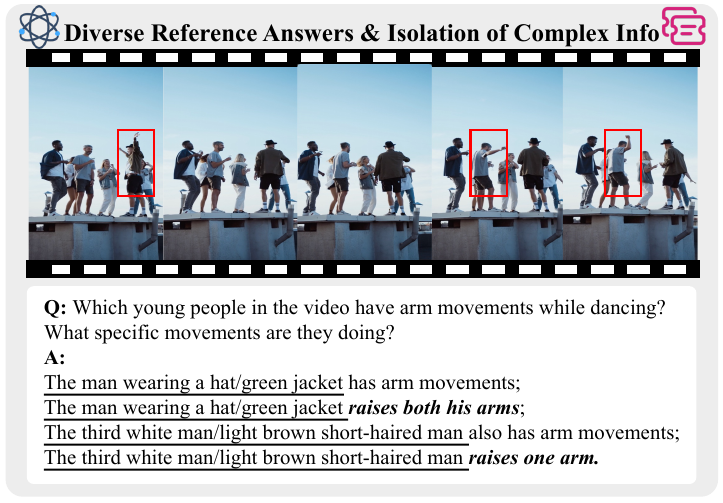}
	\caption{Separating scoring metrics: (1) presence of arm movements and (2) movement specificity, to systematically isolate complex information evaluation. Concurrently, character-specific features (e.g., wearing hat, wearing green jacket) are leveraged to formulate diverse reference answers, and therefore enhance answer adaptability across diverse caption.
}
	\label{fig:metric2}
\end{figure}
\subsection{VC4VG-Bench}
\label{cap4gen-bench}


To quantitatively evaluate caption coverage accuracy across critical video reconstruction dimensions and assess corresponding T2V generation improvements, we introduce VC4VG-Bench, an automated evaluation caption benchmark for T2V. 

\subsubsection{Evaluation Dimensions and Videos}
Aligning with the characteristics of a detailed caption necessary to generate high-quality video, our benchmark encompasses evaluations in five critical dimensions of videos mentioned in Section~\ref{method:dimensions}.
Therefore, in terms of video collection, rather than achieving diversity through disparate data sources, we prioritize the diversity of videos across the five evaluation dimensions. 
The evaluation videos are curated from Pixabay\footnote{\url{https://pixabay.com/videos}}, chosen for their high aesthetic quality and rich visual detail, with durations typically ranging from 5 to 20 seconds. 

\subsubsection{Evaluation QA Design}
In terms of evaluation QA system design, We adopt a similar divide-and-conquer strategy by AuroraCap~\cite{chai2024auroracap}.
\paragraph{Human Annotation Strategy} Unlike AuroraCap~\cite{chai2024auroracap}’s approach, which relies on manually refined ground-truth captions derived from LLM-generated outputs and fully automates QA generation using GPT-4~\cite{openai2023gpt4} with predefined prompts, our QA pairs are entirely human-annotated as shown in Figure~\ref{fig:cap4gen bench}. 
Annotators simultaneously reference both the original video content and Gemini-1.5-Pro~\cite{team2024gemini1.5} generated captions—the latter of which may contain information omissions or hallucinations. 
This dual-reference methodology creates a complementary framework where human visual interpretation and multi-modal model understanding jointly establish a holistic and precise comprehension of video content. 

We opt for manual QA annotation over manual caption refinement to ensure that our QA design incorporates diverse granularity and complexity levels to assess nuanced information reconstruction. Directly generating QA pairs by LLMs exhibits the inherent reliability limitations.

\paragraph{Temporal Information Processing} 
In terms of question formulation, temporal information introduces significant complexity, particularly when considering sequences of actions (e.g., motion trajectories of subjects or camera operations) that involve chronological ordering, concurrent events, or causal relationships. 

We address this by clustering temporally correlated information (e.g., sequences of hand movements) for evaluation. This design is motivated by two primary considerations: First, aggregating multiple temporal elements into a single question (e.g., "What sequential actions did the subject perform?") would substantially increase the difficulty of answer formulation and evaluation. Second, decomposing sequences into individual actions risks introducing conditional dependencies (e.g., "What occurred after Action 1?"), which becomes unmanageable if the caption omits or misrepresents prerequisite actions (e.g., Action 1).
\paragraph{General QA Formulation} 
To further enhance assessment robustness against variations in captioner outputs (e.g., linguistic diversity, descriptive paradigms, accuracy, comprehensiveness, and granularity), we implement three general strategies as shown in Figure~\ref{fig:metric1} and Figure~\ref{fig:metric2}: 

1) \textit{Multigranularity QA supplementation}: Incorporating questions that assess both fine-grained details (e.g., enumerating specific hand movements) and high-level assertions (e.g., presence/absence of hand actions);

2) \textit{Isolation of complex information}: Separating challenging elements (e.g., left/right hand distinctions) from broader contextual descriptions to avoid conflated evaluations;

3) \textit{Diversified reference answers}: Accommodating multiple valid descriptions for ambiguous entities (e.g., ``the man on the left'' vs. ``the man wearing a black hat'') through semantically equivalent answer variants.  


\begin{table*}[h!]
\label{caption bench result}
\centering
\resizebox{1.0\linewidth}{!}{%
\begin{tabular}{c|ccccc|cc|c}
\toprule
\multirow{2}{*}{Caption Model} & Environment & Subject & Motion & Camera & Atmosphere\&style & Necessity-L1 & Necessity-L2 &Total score \\
& Score/\% & Score/\% & Score/\% & Score/\% & Score/\% & Score/\% & Score/\%& Score/\%\\
\midrule
    ShareCaptioner-Video-7B~\cite{chen2025sharegpt4video} & 196/43.5 & 103/22.3 & 85/25.4 & 48/33.1 & 12/70.6 & 284/46.3 & 160/20.1 & 444/31.5 \\
    Vriptor~\cite{yang2024vript} & 208/46.1 & 126/27.3 & 60/17.9 & 31/21.4 & 16/94.1 & 303/49.3 & 138/17.3 & 441/31.3  \\
    VideoLLaMA3-7B~\cite{damonlpsg2025videollama3} & 119/26.4 & 106/22.9& 88/26.3 & 17/11.7 &14/82.4 & 232/37.8 & 112/14.1 & 344/24.4 \\
    Qwen2VL-7B~\cite{qwen2-VL} & 179/39.7 & 134/29& 98/29.3 & 23/15.9 &12/70.6 & 296/48.2 & 150/18.8 &446/31.6 \\
    CogVLM2-Caption~\cite{yang2024cogvideox}  & 216/47.9 & 174/37.7&93/27.8 &14/9.7 &13/76.5 & 317/51.6& 193/24.2 &510/36.2 \\
    LLaVA-Video-7B~\cite{zhang2024llavavideo} & 287/63.6 & 211/45.7& 110/32.8 &28/19.3 &15/88.2 & 367/59.8 & 284/35.7 &651/46.2 \\
    Gemini 1.5 Pro~\cite{team2024gemini1.5} & 278/61.6 & 255/55.2 & 119/35.5 & 44/30.3 & \textbf{17/100.0} & 374/60.9 & 339/42.6 & 713/50.6 \\
\midrule
    LLaVA-Video-Gen-7B(Ours) & 304/67.4 & 256/55.4 & 154/46.0 & 74/51.0 & 16/94.1 & 459/74.8 & 345/43.3 & 804/57.0 \\
    Gemini 1.5 Pro-MiraData~\cite{ju2025miradata} & \underline{335/74.3} & \underline{287/62.1} & \underline{163/48.7} & \underline{77/53.1} & 16/94.1 & \underline{471/76.7} &\underline{407/51.1}  & \underline{878/62.3} \\
    Gemini 1.5 Pro-VC4VG~\cite{team2024gemini1.5} & \textbf{372/82.5} & \textbf{328/71.0}& \textbf{170/50.7} & \textbf{85/58.6} &\textbf{17/100.0} & \textbf{513/83.6}& \textbf{459/57.7} & \textbf{972/68.9} \\
\bottomrule
\end{tabular}%
}

\caption{Quantitative captioning evaluation results comparison between free-generated and content-constrained models. The best results of video captioning methods are marked in \textbf{bold} and the second-best are \underline{underlined}. It is important to note that due to inherent differences of model and variations in prompt engineering strategies, the caption results do not reflect their absolute performance capabilities. For free-generated setting, models response using the uniform prompt \texttt{"Please describe this video in detail"}. }
\label{tab:capbencheval}
\end{table*}

\begin{figure}[!t]
	\centering
	\includegraphics[width=\linewidth]{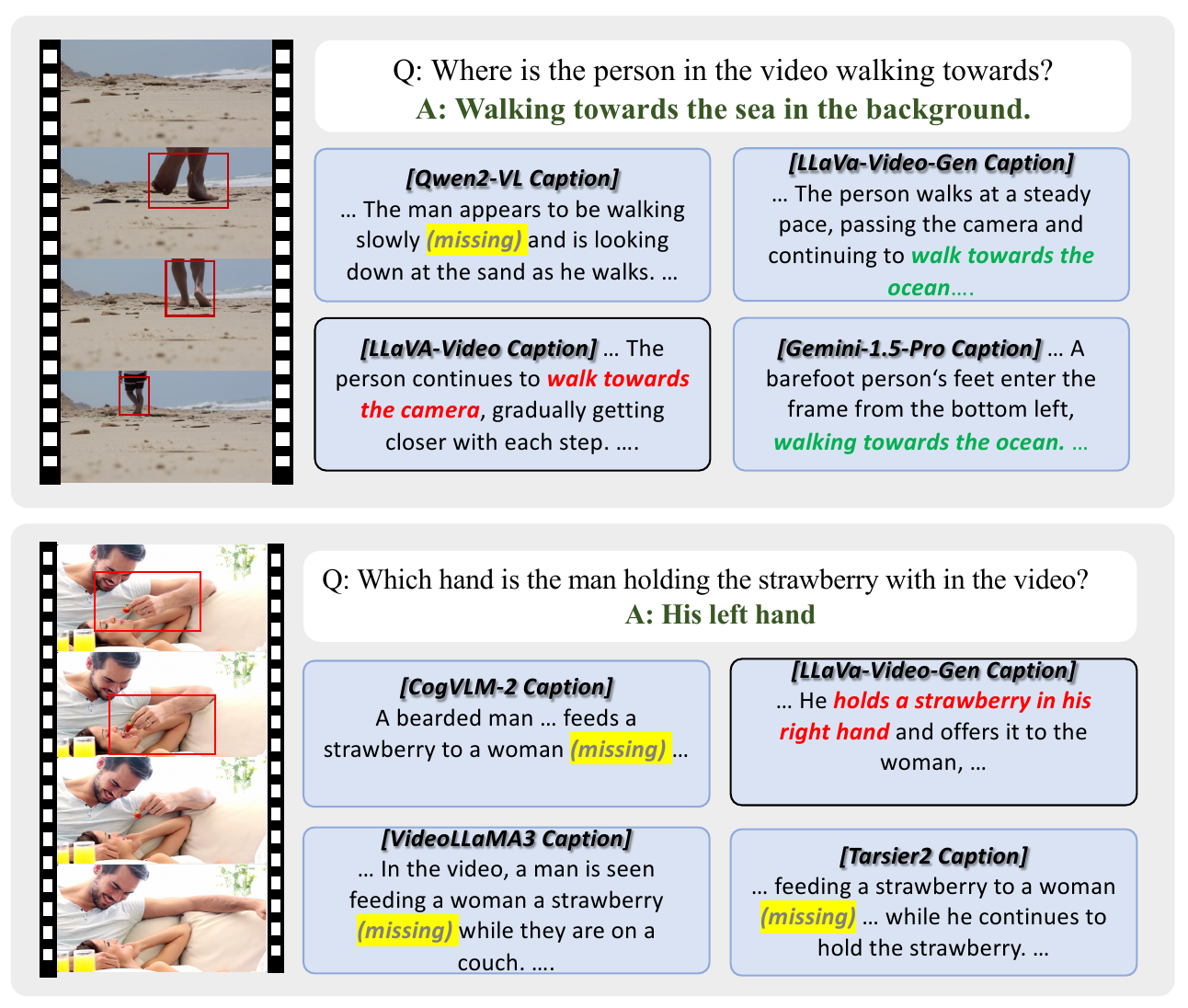}
	\caption{Illustration of representative examples of video caption performance on the benchmark, demonstrating variations in action descriptions.}
	\label{fig:capbench_case}
\end{figure}
\subsubsection{Evaluation Metrics}
In the design of evaluation metrics, we allocate scores based on the informational density of each QA pair. For QA pairs containing substantial information, we decompose answers into multiple scoring points to enable precise score distribution while reducing the complexity of automated evaluation. 

\noindent\textbf{Reconstruction-necessity-based Hierarchy.} We stratify QA pairs into two levels according to their necessity for video reconstruction. This hierarchy reflects our expectation that captions should prioritize accurate coverage of information critical to video fidelity. Regarding the classification criteria for reconstruction-necessity-based  hierarchy, information pertaining to high-level concepts and core structures is predominantly categorized as Level-1 necessity, while fine details are generally assigned to  Level-2 necessity. Concurrently, the dimension of information or its visual saliency level within the video context also impacts necessity classification. For instance, although both represent fine details, the color of the dress of the subject female (as the visual focus) would be classified as Level-1 necessity, whereas the color of background curtains (secondary visual elements) would typically fall under Level-2 necessity. 

\subsubsection{Automated Evaluation Results}
We adopt the LLM-as-judge paradigm to implement automated evaluation, leveraging GPT-4o for extracting target information from captions and determining whether predefined scoring criteria are adequately addressed. The pipeline achieved a consistency rate over 80\% with human judgments, which demonstrates the reliability of our framework. 

As demonstrated in Table~\ref{tab:capbencheval}, under the free-generated setting, mainstream MLLMs and specialized captioners exhibit significant performance variations on our benchmark. 
Gemini-1.5-Pro demonstrates relative advantages overall. However, without explicit prompt guidance, it tends to generate concise and generalized captions that frequently omit details essential for video reconstruction. 


CogVLM2-Caption~\cite{yang2024cogvideox}, ShareCaptioner-Video-7B~\cite{chen2025sharegpt4video} and Vriptor~\cite{yang2024vript}, despite being specialized captioning models, exhibit deficiencies across multiple dimensions and therefore struggle to generate captions that effectively support text-to-video applications.



Under the prompt engineering setting, we compared two data synthesis strategies for T2V tasks, MiraData~\cite{ju2025miradata} and our VC4VG, using Gemini-1.5-Pro. Both approaches emphasize comprehensive descriptions across video dimensions, where the former requires structured caption output while the latter imposes no format restrictions. Benchmark results demonstrate that Gemini-1.5-Pro-VC4VG achieves significantly higher scores than Gemini-1.5-Pro-MiraData, which in turn significantly outperforms Gemini-1.5-Pro under free-generated setting. This suggests that while MiraData's synthesis strategy can effectively align with critical dimensions of T2V tasks, there remains room for improvement. 

Our captioning model trained on Gemini-1.5-Pro-VC4VG data demonstrates competitive performance on the benchmark. Compared to Gemini-1.5-Pro under free-generated setting, it shows significant improvements at the primary necessity-level, approaching the performance level of Gemini-1.5-Pro-MiraData. This indicates that the captions generated by our model can accurately and comprehensively describe the highly essential information across various dimensions required for video reconstruction.

\section{T2V Generation Experiments}

\begin{table*}[h!]
\setlength\tabcolsep{3pt}
\centering
\resizebox{1.0\linewidth}{!}{%
\begin{tabular}{p{4.5cm}|c|c|c|c|c|c|c|c}
\toprule
\multirow{2}{*}{Captioning Models} & Subject & Background & Temporal & Motion & Dynamic & Aesthetic & Imaging & Object\\
& Consistency & Consistency & Flickering & Smoothness & Degree & Quality&Quality&Class\\
\midrule
    CogVideoX-5B & 92.93\% & 94.41\%&97.95\% &97.76\%&\textbf{68.06\%} &61.93\% &61.26\% &82.20\%\\
    ~+CogVLM2-Caption  & 93.60\% & 95.31\%& 95.45\% &98.73\% &58.33\% &63.43\% &64.02\%&88.37\%\\
    ~+LLaVA-Video-7B & 93.59\% & 95.12\%&\textbf{98.53\%} &\textbf{98.79\%} &59.72\% & 64.00\% &63.47\%&87.74\%\\
    ~+LLaVA-Video-Gen(Ours) & \textbf{94.25\%} & \textbf{95.58\%}&98.20\% &98.56\% &59.72\% & \textbf{65.16\%} &\textbf{65.95\%}& \textbf{90.98\%}\\
\bottomrule
\toprule
\multirow{2}{*}{Captioning Models} & Multiple & \multirow{2}{*}{Color} & Spatial & \multirow{2}{*}{Scene} & Temporal & Appearance & Overall & Total\\
& Objects &   & Relationship &   & Style & Style&Consistency&Score\\
\midrule
    CogVideoX-5B & 57.62\% & 78.63\%&60.66\% &51.67\% &24.95\% &23.99\% &27.07\%&79.97\%\\
    ~+CogVLM2-Caption  & 63.33\% & 79.58\% &73.45\% &56.32\% &25.60\% &\textbf{24.68\%} &27.55\%&81.54\%\\
    ~+LLaVA-Video-7B & 70.88\% & \textbf{85.21\%}&71.37\% &53.85\% &\textbf{25.78\%} &24.16\% &27.59\%&81.79\% \\
    ~+LLaVA-Video-Gen(Ours) & \textbf{77.90\%} & 75.84\%&\textbf{75.65\%} &\textbf{59.88\%} &25.64\% &24.56\% &\textbf{27.70\%}&\textbf{82.50\%}\\
\bottomrule
\end{tabular}%
}
\caption{Quantitative VBench evaluation results comparison between T2V models trained with captions generated by different models. We use all dimension gpt enhanced prompts in vbench and sample once for each prompt. The best results of video captioning methods are marked in \textbf{bold}.}
\label{tab:vbenchresult}
\end{table*}

\begin{table*}[h!]
\centering
\small
\resizebox{1.0\linewidth}{!}{%
\begin{tabular}{c|ccccc|c}
\toprule
\multirow{2}{*}{Captioning Models} & Environment & Subject & Motion & Camera & Atmosphere\&style & Overall\\
& G/S/B/\% & G/S/B/\% & G/S/B/\% & G/S/B/\% & G/S/B/\% & G/S/B/\% \\
\midrule
    LLaVA-Video-Gen&-&-&-&-&-&- \\
    \textit{-vs} LLaVA-Video-7B&26.5/72/1.5&50/44/6&23.5/68.5/8&0.5/98.5/1&1/99/0&61/28.5/10.5 \\
    \textit{-vs} CogVLM2-Caption&16/82.5/1.5&28.5/62.5/9&23.5/68.5/8&1/97.5/1.5&0/99.5/0.5&37.5/51/11.5 \\
\bottomrule
\end{tabular}%
}
\caption{Quantitative human-annotated evaluation results. The evaluation compares the performance of LLaVA-Video-Gen, against two baseline models: LLaVA-Video-7B and CogVLM2-Caption. Human annotators assessed video outputs from these models based on 200 samples from the MovieGenBench dataset, which are annotated with prompts in miradata-style~\cite{ju2025miradata} For each comparison, evaluators rated whether LLaVA-Video-Gen's output was Good (G), Same (S), or Bad (B) relative to the baseline across several criteria. The scores are presented as G:S:B percentages, indicating the proportion of times LLaVA-Video-Gen is judged superior, equivalent, or inferior to the respective baseline for each dimension.}
\label{tab:moviegen_result}
\end{table*}

In this section, we present experimental results and analysis of applying different captioning methods to CogVideoX-5B~\cite{yang2024cogvideox} T2V model training. Section~\ref{training_prepare} details our training preparation including video sources, captioning methodologies, and parameter configurations. We subsequently demonstrate the effectiveness of video-caption pairs generated by different captioning models for T2V model training in Section~\ref{sec:t2v_results}.
\subsection{Experimental Settings}
\label{training_prepare}
\noindent\textbf{Video Source and Preprocessing}:
We curated approximately 72K videos from OpenVid-1M~\cite{nan2024openvid} through rigorous filtering based on aesthetic quality and temporal consistency. To mitigate aspect ratio distortion caused by resolution mismatches during training, we implement adaptive resizing and cropping based on each video's original aspect ratio. Given that CogVideoX-5B generates 6-second videos with 49 frames at 8 frames per second (fps), we temporally segment all source videos into 6-second clips through random sampling to ensure motion consistency. This refined dataset serves as our primary video source for validating different captioning methodologies.

\noindent\textbf{Captioning Methods}:
Consistent with the captioning guidelines in Table~\ref{tab:capbencheval}, we employ the following models for video caption generation:
(1)CogVLM2-Caption~\cite{yang2024cogvideox} is adopted during the training of CogVideoX to convert video data into textual descriptions. This alignment tends to ensure consistency between the fine-tuning phase and CogVideoX's training paradigm.
(2)LLaVA-Video-7B~\cite{zhang2024llavavideo} extends the LLaVA-Onevision ~\cite{li2024onevision} through fine-tuning on the LLaVA-Video-178K which containing detailed caption annotations, enabling the generation of comprehensive and fine-grained video descriptions.
(3)LLaVA-Video-Gen represents our expert captioner model introduced in Section~\ref{method:dimensions}, which is distilled from Gemini 1.5 Pro with prompt enhanced on dimensions mentioned in Sec~\ref{method:dimensions}.

\noindent\textbf{T2V Model Setting}:
We conduct full-parameter fine-tuning of CogVideoX-5B, a widely adopted open-source DiT-based T2V generation model, using the original training configuration: 49-frame sampling, 720×480 resolution, learning rate of 2e-5, and 64×NVIDIA H20 GPUs for 5 epochs.
During inference, we maintain identical resolution and frame count as in training, configured with 8 fps to generate approximately 6-second videos. The CogVideoXDPMScheduler~\cite{lu2022dpm, lu2022dpm++} is employed with 50 steps and guidance of scale 6 throughout inference phases.
\subsection{Experimental Results Comparison}
\label{sec:t2v_results}

\subsubsection{Automatic Quantitative Evaluation}
\paragraph{Automatic Metrics.} We employ several metrics in VBench~\cite{huang2023vbench}, a widely adopted benchmark for automated evaluation of T2V generation quality, to assess models trained with different captioning methods. Given that our training utilizes extended captions containing richer visual details and motion descriptions, we adopt the official GPT-enhanced prompts from VBench repository for generation. As shown in Table~\ref{tab:vbenchresult}, LLaVA-Video-Gen demonstrates superior overall performance in most of the metrics, especially for semantic understanding such as multiple objects, spatial relationship and scene. The performance ranking aligns with our VC4VG-Bench scores from Table~\ref{tab:capbencheval}, validating our benchmark's effectiveness for evaluating training captions.

\subsubsection{Human-annotated GSB Quantitative Evaluation}
To enable fine-grained evaluation of T2V generation fidelity, we curate 200 samples from MovieGenBench~\cite{polyak2024moviegencastmedia}. Using Gemini-1.5-Pro, we generate Miradata-style prompts with MovieGen-produced videos as reference, then reconstruct videos through each T2V model. Three domain experts perform blind assessments comparing LLaVA-Video-Gen against its closest-performing counterparts (LLaVA-Video-7B and CogVLM-Caption) through side-by-side evaluation using GSB (Good, Same, Bad) scoring criteria across five reconstruction dimensions.

Our findings in Table~\ref{tab:moviegen_result} reveal three key insights: (1) Information gains in Environment, Subject, and Motion dimensions directly correlate with T2V generation improvements; (2) Comparable performance on Atmosphere attributes across models aligns with VC4VG-Bench's lower task difficulty for this dimension; (3) For Camera properties, while models effectively control shot size and angles, movement patterns prove challenging due to MLLMs' limited capability in understanding fine-grained temporal dynamics - a limitation exacerbated by MovieGenBench's sparse coverage of complex camera motions.
Collectively, these empirical results validate that our dimension-aware optimization strategy effectively guides T2V training data curation.

\subsubsection{Qualitative Evaluation}
\begin{figure}[!t]
	\centering
	\includegraphics[width=\linewidth]{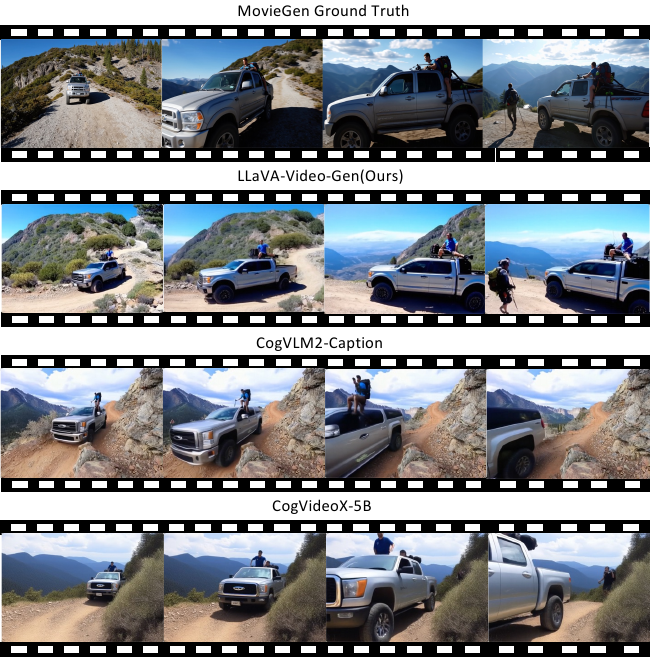}
	\caption{Qualitative evaluation of different T2V models' reconstruction performance. Please zoom in for a better view.}
	\label{fig:moviegenbench_case}
\end{figure}
We choose samples for Figure~\ref{fig:moviegenbench_case} and Figure~\ref{fig:t2v_qualitative} to visualize representative cases. The T2V model fine-tuned on captions generated by different models demonstrates T2V improvements in scene detail preservation and instruction adherence compared to the raw CogVideoX-5B.


\section{Related Works}

\noindent\textbf{Video-Text Dataset.}
High-quality T2V models require video-text datasets with scene details and instruction alignment for effective training. Existing datasets primarily fall into three categories: human-annotated~\cite{xu2016msrvtt, Du2024RTime, wang2019vatex, anne2017didemo}, metadata-derived captions from video platforms~\cite{Bain21webvid}, and automatically generated captions~\cite{miech19howto100m, chen2024panda70m, wang2023internvid, yang2024vript, nan2024openvid, ju2025miradata}. 
Traditional automation methods like ASR transcription~\cite{miech19howto100m, xue2022hdvila} achieve scale but exhibit weak video-text semantic alignment, making them suboptimal for generative tasks.

Modern multimodal LLMs (MLLMs) demonstrate enhanced visual description capabilities, driving their adoption in T2V training corpus generation~\cite{chen2024panda70m, wang2023internvid, nan2024openvid, opensora, hong2022cogvideo, yang2024cogvideox,kong2024hunyuanvideo, polyak2024moviegencastmedia, ju2025miradata, chen2025sharegpt4video, yang2024vript}. Datasets like Panda-70M~\cite{chen2024panda70m} and InternVid~\cite{wang2023internvid} only produce short captions. Current solutions prioritize fine-grained dense video descriptions through MLLM-based approaches: OpenSora~\cite{opensora} leverages PLLaVA~\cite{xu2024pllava}, CogVideoX~\cite{yang2024cogvideox, hong2022cogvideo} employs its proprietary CogVLM2-Cap, OpenVid utilizes LLaVA-1.6~
\cite{liu2024llavanext}, and MiraData~\cite{ju2025miradata} adopts cost-intensive GPT-4V~\cite{zhang2023gpt4vision} annotations. Most methods adopt approaches without specialized frameworks for optimizing video generation elements. InstanceCap~\cite{fan2024instancecap} generates dense structural captions through a complex pipeline and suffers from significant efficiency bottlenecks compared to end-to-end generation methods, ultimately limiting its scalability.

\noindent\textbf{Evaluation of Video Captioning.} 
As the capabilities of video captioning have advanced, the associated benchmarks have evolved from traditional short-caption evaluation(e.g., MSR-VTT~\cite{xu2016msrvtt}, VATEX~\cite{wang2019vatex}) and metrics(e.g., METEOR~\cite{banerjee2005meteor} CIDEr~\cite{vedantam2015cider}, BLEU~\cite{papineni2002bleu}, ROUGE-L~\cite{lin2004rouge}), to address long-form captioning challenges.
Notably, AuroraCap~\cite{chai2024auroracap} introduced VDC~\cite{chai2024auroracap}, along with an LLM-based evaluation metric VDCScore, overcoming limitations of direct caption assessment through LLMs. 
Dream-1K~\cite{wang2024tarsier_recipestrainingevaluating} and CaReBench~\cite{xu2025carebenchfinegrainedbenchmarkvideo} focus more extensively on human-annotated video captions and tailored evaluation methods. However, these benchmarks are primarily designed for video captioning in the context of video understanding rather than video generation. 
Although VidCapBench~\cite{chen2025vidcapbench} aligns its evaluation design with the key metrics for T2V generation, its training-free T2V verification mechanism inadequately demonstrates that models performing well on this benchmark can effectively serve as training data for high-quality T2V generation. In this paper, we propose a novel benchmark specifically designed for T2V tasks and empirically validate its consistency with actual generation quality through real-world T2V training experiments.



\section{Conclusion}

In this paper, we present VC4VG, a comprehensive video caption optimization framework designed for T2V models. Our framework systematically decomposes video captioning into five key dimensions that are critical for video reconstruction, thereby providing practical guidance for scalable caption generation. Building on this decomposition, we further introduce VC4VG-Bench, a specialized benchmark that emphasizes multi-dimensional video descriptions tailored to T2V generation scenarios. Through fine-tuning experiments, we demonstrate a clear correlation between enhanced caption quality and improved video generation performance, validating the effectiveness of our approach. We hope that our framework will support the community in developing higher-quality video captions for T2V models and, ultimately, more powerful video generation systems.

\section*{Limitations}

Our VC4VG-Bench automates the evaluation of open-ended video captioning. While demonstrating high correlation with human judgment, subtle biases may still exist. Furthermore, performance can fluctuate due to varying model configurations, including different video processing techniques and prompt engineering strategies. Consequently, the reported metrics primarily reflect caption quality under specific experimental settings, rather than the fundamental performance differences between the models.

\section*{Ethical Considerations}

Regarding ethical considerations, it is important to acknowledge that Text-to-Video models may generate biased or harmful content. Such outputs can potentially perpetuate stereotypes or disseminate misinformation. We emphasize the critical need for responsible model application. Developers are encouraged to implement robust safeguards to mitigate these risks.

 \section*{Acknowledgments}

 This work was sponsored by CCF-ALIMAMA TECH Kangaroo Fund (NO. CCF-ALIMAMA OF 2024007).


\bibliography{custom}

\appendix

\label{sec:appendix}

\begin{figure*}[h!]
	\centering
	\includegraphics[width=\linewidth]{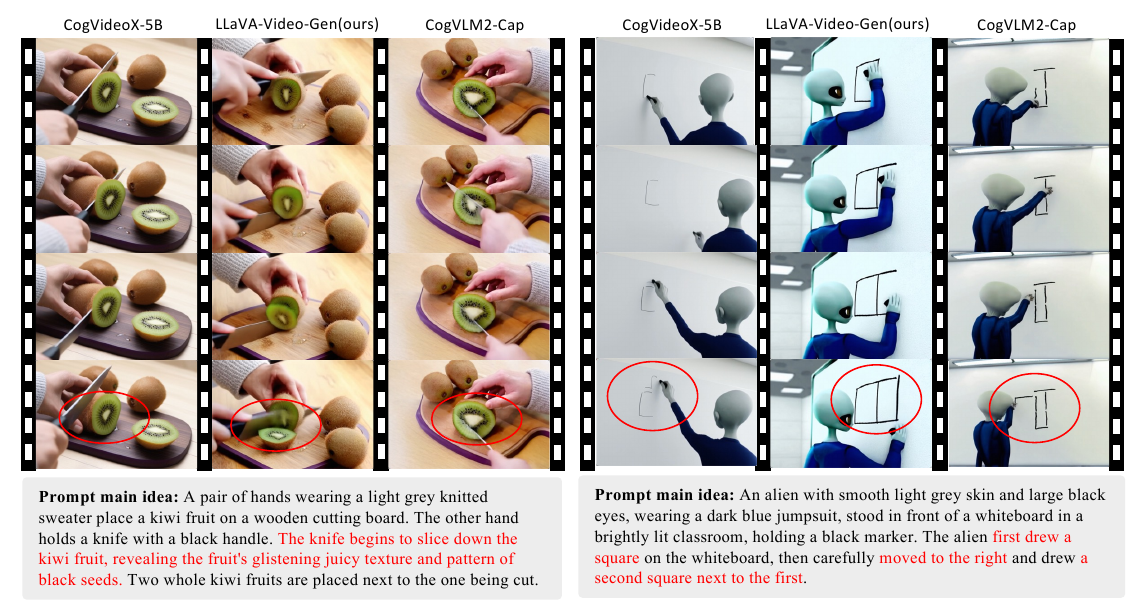}
	\caption{Qualitative comparison of CogVideoX-5B between raw checkpoint and versions trained on captions generated by LLaVA-Video-Gen and CogVLM2-Cap. Due to space limitations, only the main idea of the prompt is shown. The red circles highlight the main distinguishing points of the generated videos. Please zoom in for a better view.
    }
	\label{fig:t2v_qualitative}
\end{figure*}
\section{Video Filtering Details}
\label{appendix:video_filter}
We implemented a proprietary data cleaning pipeline to rigorously process the OpenVid-1M~\cite{nan2024openvid} dataset, ultimately curating 72K high-quality videos. The pipeline integrates the following critical components:
\begin{itemize}
\item Text Overlay Detection: Detects excessive subtitles or text overlays in videos, filtering out frames with significant content obstruction.

\item Aesthetic Score and DOVER++~\cite{wu2023dover}: Evaluates visual quality by sampling multiple frames per video clip, applying the DOVER++ assesses overall video quality, considering technical and aesthetic factors, to discard low-quality videos.

\item Video Classification \& Frame-level Filtering: we developed a classification model to detect low-quality content categories, including frosted-border videos and PPT-style slideshows.
We filter videos with transitional effects (e.g., fade-in/fade-out) through per-frame analysis to ensure content consistency.

\item Optical Flow-based Motion Intensity Resampling: Utilizes the RAFT~\cite{teed2020raft} model to compute optical flow from video frames, quantifying motion intensity distribution to guide training data resampling.
\end{itemize}

\begin{table}[t]
\centering
\resizebox{\linewidth}{!}{%
\begin{tabular}{c|ccc}
\toprule
\multirow{1}{*}{Satistics} & QA Pair & Scoring Point & Avg Point/Pair \\
\midrule
    Subject  &  293	& 462	& 1.6 \\
    Environment & 306 &	450	& 1.5 \\
    Atmosphere\&Style & 17	&17	&1.0 \\
    Motion & 208	&335	&1.6\\
    Camera Info & 132	&145	&1.1\\
\midrule
    Necessity-L1 & /	&614	&/ \\
    Necessity-L2 & /	&796	&/\\
\midrule
    Total & 956	&1410	&1.5\\
\bottomrule
\end{tabular}%
}
\caption{VC4VG-Bench Statistics.}
\label{tab:capbenchstatistics}
\end{table}

\section{VC4VG-Bench Details}
\subsection{Statistics}
We show the statistics of VC4VG-Bench in Table~\ref{tab:capbenchstatistics}.
As illustrated, the distribution of QA pairs across different dimensions is intentionally non-uniform. This design choice reflects the relative importance and information richness of each dimension in the context of text-to-video generation.

Specifically, dimensions such as \textit{Subject}, \textit{Environment}, and \textit{Motion} are fundamental to producing coherent and meaningful video content. They typically carry the majority of the semantic information within a video description. Therefore, we allocate a larger portion of our annotation budget to these core dimensions, resulting in a higher number of QA pairs to ensure comprehensive evaluation of these essential aspects.

In contrast, the \textit{Atmosphere \& Style} dimension, while important for personalization and artistic expression, is often more subjective and can be described with fewer words. To maintain a high standard of objectivity in our benchmark, we adopt a conservative annotation strategy for this category. We create QA pairs for \textit{Atmosphere \& Style} only when such attributes are visually distinct and could be described objectively. For videos that lack a clear and discernible style, no QA pair is assigned for this dimension. This deliberate approach explains why \textit{Atmosphere \& Style} has the fewest QA pairs. This rationale ensures that our benchmark effectively prioritizes the most critical and objectively measurable aspects of video generation.

\subsection{Prompt Template}
\begin{figure}[!t]
  \centering
	\includegraphics[width=\linewidth]{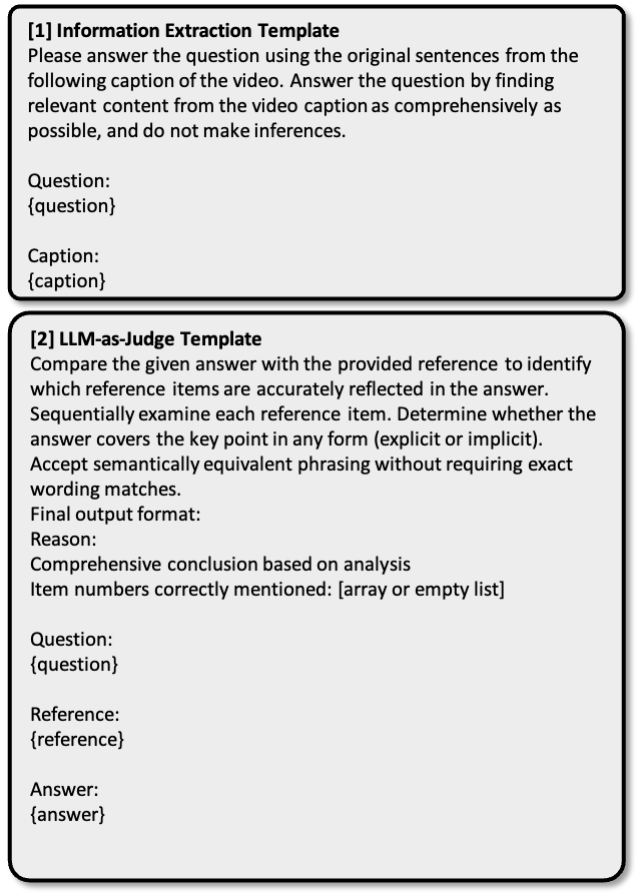}
	\caption{Automated Evaluation Prompt Template}
	\label{fig:prompt_template}
\end{figure}

In the automated evaluation process, we first extract question-relevant content from the generated captions, then assess the extracted information by comparing it with reference answers. The corresponding prompt template for this evaluation pipeline is demonstrated in Figure~\ref{fig:prompt_template}. We employ GPT-4o-0806 version as the evaluation judge, utilizing its reasoning capabilities to perform content alignment analysis and scoring. 
\subsection{Video Collection}
\begin{figure*}[!t]
  \centering
	\includegraphics[width=0.5\linewidth]{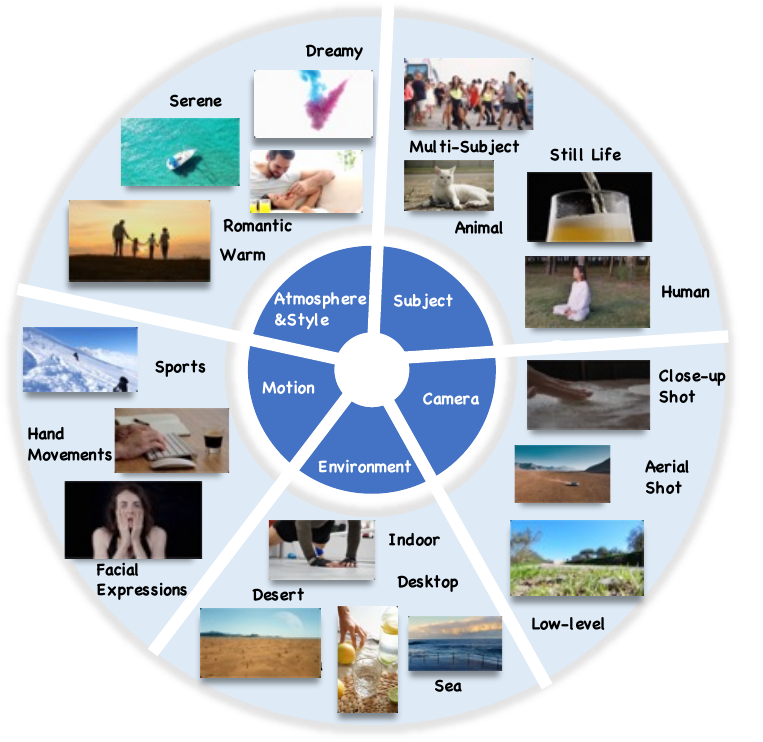}
	\caption{Video Examples from Benchmark}
	\label{fig:video_example}
\end{figure*}

Video selection was primarily based on diversity across caption dimensions, which inherently ensures content diversity in the visual domain.Figure~\ref{fig:video_example} presents video examples from our benchmark, demonstrating the corresponding video diversity across various dimensions.

\begin{table*}[ht]
\centering
\resizebox{\linewidth}{!}{%
\begin{tabular}{lcccccc}
\toprule
\multirow{2}{*}{Caption Model} & Environment & Subject & Motion & Camera & Atmosphere\&style & Total score\\
& Score/\% & Score/\% & Score/\% & Score/\% & Score/\% & score/\%\\
\midrule
LLaVA-Video-7B (Baseline) & 287/63.6 & 211/45.7 & 110/32.8 & 28/19.3 & 15/88.2 & 651/46.2 \\
LLaVA-Video-PE (w/o fine-tuning) & 240/53.2 & 183/39.6 & 117/34.9 & 44/30.3 & 15/88.2 & 599/42.5 \\
LLaVA-Video-Gen-SFT (w/o RTime) & 289/64.1 & \textbf{258/55.8} & 146/43.6 & 71/49.0 & \textbf{16/94.1} & 780/55.3 \\
\textbf{LLaVA-Video-Gen} (Final) & \textbf{304/67.4} & 256/55.4 & \textbf{154/46.0} & \textbf{74/51.0} & \textbf{16/94.1} & \textbf{804/57.0} \\
\bottomrule
\end{tabular}
}

\caption{Ablation study of LLaVA-Video-Gen on the VC4VG-Bench. We evaluate the impact of our curated WebVid data (SFT) and the RTime dataset (DPO). 'PE' denotes Prompt Engineering. Best results are in \textbf{bold}.}
\label{tab:ablation_gen}
\end{table*}

\section{Other Experiments Details}

\subsection{Ablation Study of LLaVA-Video-Gen}
\label{appendix:abalation_llava}
To validate the effectiveness of our training strategy and dissect the contributions of each component, we conduct a thorough ablation study for LLaVA-Video-Gen on the VC4VG-Bench. We evaluate four distinct model variants to isolate the impact of our data curation and fine-tuning methods. The variants are as follows:
\begin{itemize}
\item \textbf{LLaVA-Video-7B}: The original pre-trained model, evaluated with simple prompts as a baseline (equivalent to our reporting in Table 2).
\item \textbf{LLaVA-Video-PE}: The original model without any fine-tuning, but prompted with our complex, multi-dimensional instructions to assess the impact of prompt engineering alone.
\item \textbf{LLaVA-Video-Gen-SFT}: The model after SFT on our curated 200K WebVid subset, but without the subsequent temporal enhancement stage.
\item \textbf{LLaVA-Video-Gen}: Our final model, which undergoes both SFT on the WebVid data and DPO on the RTime dataset.
\end{itemize}
The results, presented in Table~\ref{tab:ablation_gen}, clearly demonstrate the efficacy of our methods. The results confirm that 1) Prompt engineering alone offers limited improvement and may even degrade performance; 2) SFT on our high-quality WebVid subset leads to substantial gains; 3) DPO with RTime yields additional improvements, especially in motion and camera dimensions.

\subsection{Ablation Study of T2V Training Steps}

\begin{figure}[!t]
  \centering
	\includegraphics[width=\linewidth]{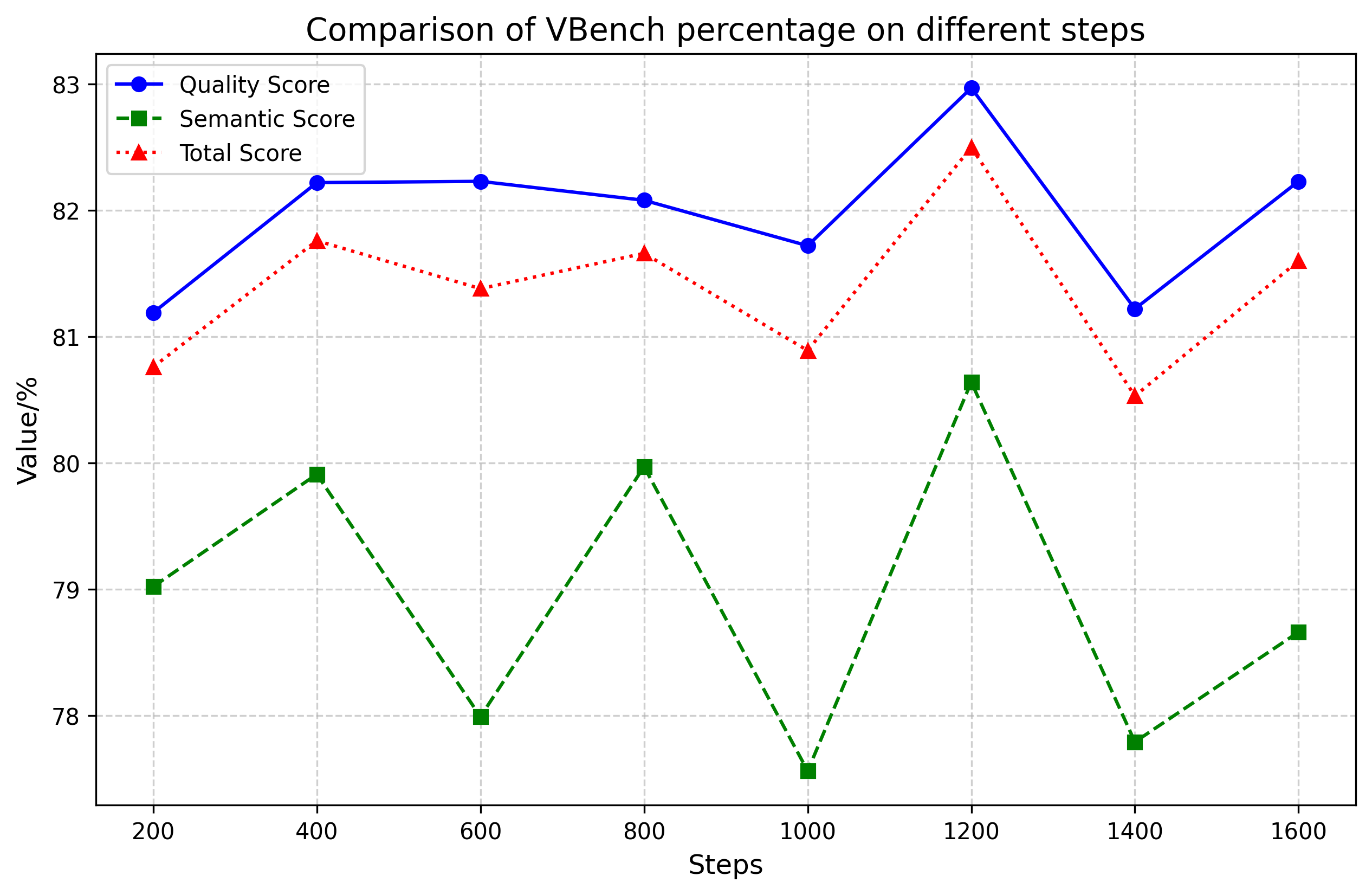}
	\caption{Comparison of VBench score percentage on different steps.}
	\label{fig:step}
\end{figure}

As illustrated in Figure~\ref{fig:step}, we fine-tune CogVideoX-5B for 5 epochs (1,600 steps) using captions generated by our LLaVA-Video-Gen framework. Based on VBench evaluations~\cite{huang2023vbench}, which measure quality score, semantic score, and total score through line chart analysis, we observe peak performance at 1,200 training steps. We therefore select the 1200-step checkpoint for final evaluation. To ensure fair comparison in Section 4.2, all baseline caption methods are evaluated under identical training configurations using their respective 1200-step checkpoints.






\subsection{Qualitative Analysis}
We present a qualitative comparison between our LLaVA-Video-Gen and CogVLM2-Caption in Figure~\ref{fig:t2v_qualitative}. 

\section{Reproducibility Statement}

We will release our benchmark and corresponding codes for reproducibility.

\section{License}
This work is licensed under the Creative Commons Attribution-NonCommercial 4.0 International License (CC BY-NC 4.0).


\end{document}